\documentclass[conference]{IEEEtran}
\IEEEoverridecommandlockouts
\pdfoutput=1
\usepackage{cite}
\usepackage{amsmath,amssymb,amsfonts}
\usepackage{algorithmic}
\usepackage{graphicx}
\usepackage{textcomp}
\usepackage{xcolor}
\usepackage{url,times}
\usepackage{color}
\usepackage{hyperref}
\usepackage{multirow, boldline}
\usepackage{xcolor, soul}

\usepackage{subcaption} 

\usepackage{pdfpages}   

\usepackage[export]{adjustbox}

\usepackage{esvect} 

\def\BibTeX{{\rm B\kern-.05em{\sc i\kern-.025em b}\kern-.08em
    T\kern-.1667em\lower.7ex\hbox{E}\kern-.125emX}}
\begin{document}


\title{Smooth Trajectory Collision Avoidance through Deep
  Reinforcement Learning}

\author{\IEEEauthorblockN{Sirui Song, \  Kirk Saunders, \  Ye Yue, \  Jundong
    Liu$^*$}
\\
  \IEEEauthorblockA{School of Electrical Engineering and
    Computer Science, \\ Ohio University,  Athens, OH 45701}
  {\thanks{$^*$Corresponding author: Dr. Jundong Liu. Email:
      liuj1@ohio.edu.  This project is supported in part by the Ohio
      University OURC program.}}}


\maketitle

\begin{abstract}

Collision avoidance is a crucial task in vision-guided autonomous
navigation. Solutions based on deep reinforcement learning (DRL) has
become increasingly popular. In this work, we proposed several novel
agent state and reward function designs to tackle two critical issues
in DRL-based navigation solutions: 1) smoothness of the trained flight
trajectories; and 2) model generalization to handle unseen
environments.

Formulated under a DRL framework, our model relies on margin reward
and smoothness constraints to ensure UAVs fly smoothly while greatly
reducing the chance of collision.  The proposed smoothness reward
minimizes a combination of first-order and second-order derivatives of
flight trajectories, which can also drive the points to be evenly
distributed, leading to stable flight speed. To enhance the agent’s
capability of handling new unseen environments, two practical setups
are proposed to improve the invariance of both the state and reward
function when deploying in different scenes. Experiments demonstrate
the effectiveness of our overall design and individual components.

\end{abstract}



\section{Introduction}





Autonomous navigation capability is of great importance for unmanned
aerial vehicles (UAVs) to fly in complex environments where
communication might be limited. Collision avoidance (CA) is among the
most crucial components of high-performance autonomy and thus has been
extensively studied.
Generally speaking, the existing CA solutions can be grouped into two
categories: geometry-based and learning-based
solutions. Geometry-based solutions are commonly formulated as a
two-step procedure: first to detect obstacles and estimate the
geometry surrounding a UAV, followed by a path planning step to
identify a traversable route for escape maneuver.


Learning-based CA solutions extract patterns from training data to
perceive environments and make maneuver decisions. Such solutions can
be broadly divided into two categories: supervised learning-based and
reinforcement learning-based. The former performs perception and
decision-making simultaneously, predicting control policies directly
from raw input images\cite{kim2015deep,giusti2015machine, tai2016deep,
  dai2020automatic, back2020autonomous}.
%
Supervised-based methods are straightforward, but they normally
require a large amount of labeled training samples, which are often
difficult or expensive to obtain.
Reinforcement learning \cite{michels2005high}, 
on the other hand, relies on a scale reward function to motivate the 
learning agent and explores policy through trial and error.
Combined with neural networks, deep reinforcement learning (DRL) has
been shown to achieve superhuman performance on a number of games by
fully exploring raw images \cite{mnih2015human, xie2017towards,
  he2020integrated}.  DRL-based collision avoidance has also been
recently proposed \cite{singla2019memory} \cite{xue2021vision}
\cite{9696380} \cite{ouahouah2021deep}.
In order to reduce cost and increase effectiveness, such training is
often first carried out within a certain simulation environment.

%
%


While remarkable progress has been made in DRL-based navigation
solutions,
insufficient attention has been given to two critical issues: 1)
smoothness of the
navigation trajectories; and 2) model generalization to handle unseen
environments.
For the former,
Kahn {\it et al}. \cite{kahn2017uncertainty} proposed a RL-based
solution that seeks a tradeoff of collision uncertainty and speed of
UAV motion. When collision uncertainty is high,
the motion of the robot/UAV is set to slower,
and vice versa. The smoothness of the flight trajectories, however, is
not directly addressed. Hasanzade {\it et
  al}. \cite{hasanzade2022deep} proposed an RL-based UAV navigation
solution based on a trajectory re-planning algorithm, where high order
B-splines are used to define and specify flight trajectories.  Due to
the local support property of B-spline, such trajectories can be
updated quickly, allowing the small UAVs to navigate in clutter
environments aggressively.  However, new knots need to be inserted over
the training process for the re-planning procedure to be fully
realized, negatively impacting the overall trajectory smoothness.

 Model generalization is a critical issue in machine learning,
 especially for DRL solutions. Many current DRL works, however, were
 evaluated on the same environments as they were trained on, such as
 Atari \cite{bellemare2013arcade}, MuJoCo \cite{todorov2012mujoco} and
 OpenAI Gym \cite{brockman2016openai}.
 For UAV training, there is an additional sim-to-real layer, which complicates
 the problem even more. 
Kong {\it et al}. \cite{yu2020path} explored the generalization of various
DRL algorithm by training them with different (but not unseen)
environments.  Doukui {\it et al}. \cite{doukhi2021deep} tackle this issue
by mapping exteroceptive sensors, robot state, and goal information to
continuous velocity control inputs, but their exploration was only
tested on unseen targets instead of unseen scenes.

  In this work, we address the afore-mentioned issues with novel
  designs for agent state and reward functions.
To ensure the smoothness of the learned flight trajectories, we
integrate two curve smoothness terms, based on first-order and
second-order derivatives respectively,
into the agent reward functions. To improve the agent’s
generalization capability, two practical setups, {\it shallow depth}
and {\it unit vector towards the target}, are adopted to boost the
robustness
of the state and reward function in dealing
with new environments.  The proposed designs are trained and tested in
simulation scenes with large geometric obstacles. 
%
Experiments 
demonstrate the effectiveness of our overall design and individual
components.

%



\section{Method}

In this work, a multirotor UAV takes off at a designated starting
point and navigates autonomously towards a destination. The line
segment connecting the start and end points is regarded as a
predefined path, along which certain objects have been put as
obstacles.


\subsection{Design and environment setup}

Our overall design goal is to fly the UAV mostly along the predefined
route while being able to avoid the obstacles. Such capability is
trained through DRL with the following considerations.
 Firstly, to ensure the UAV to fly along the predetermined path, we
 minimize the distance of the drone’s trajectory away from such a
 path. Secondly, to ensure the drone avoids collisions while flying
 smoothly, we set up a variety of rewards, including those for {\it
   margin}, {\it arrival}, and {\it penalty for collisions} and {\it
   smoothness}. In addition, we aim to design DRL agent with a good
 generalization capability in handling unseen environments.

\begin{figure}[t]
\centering
  \begin{tabular}{cc}
  {\includegraphics[width=4cm]{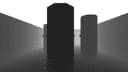}} &
  {\includegraphics[width=4cm]{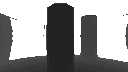}} \\ 
  (a): Original {\it deep depth}  & (b):  Truncated {\it shallow depth} 
    \end{tabular}
  \caption{An example pair of deep and shallow depth maps.}
  \label{fig:depth}
\end{figure}

{\it States}, {\it actions}, and {\it rewards} are three basic
components of most DRL algorithms.
%
%
In this work, the state $s_t$ at time $t$ is defined to include three
components: 1) the depth map of the current view facing the camera; 2)
the current velocity of the UAV, and 3) a unit vector pointing from
the UAV's current position to the target.

Our choices of 1) and 3) are both with model generalization in mind.  At
each time point, a depth image is obtained from the onboard monocular
camera of the UAV.  In order to limit the impact of environment
changes and thus improve the generalization capability of our model,
we focus on nearby objects and ignore those beyond a certain distance.
We call this truncated depth image as {\it shallow depth}, in contrast
with the original {\it deep depth}. An example pair are shown in
Fig~\ref{fig:depth}.  We include {\it unit vector to the target} as
part of the agent's state, which later will also be used in our
proposed reward function. This is in contrast to the Euclidean
distance of the UAV away from the destination. Compared with the
distances, our unit vectors are scale invariant, and therefore have
better generalization potentials to deal with new environments of
different sizes.
Each action $a_t$ at time $t$ is defined as $(v_{x_t}, v_{y_t})$, a
velocity vector with x-axis and y-axis components. The proposed reward
functions will be explained in the next subsection.



\begin{figure*}
	\centering
	\includegraphics[width=0.9\textwidth]{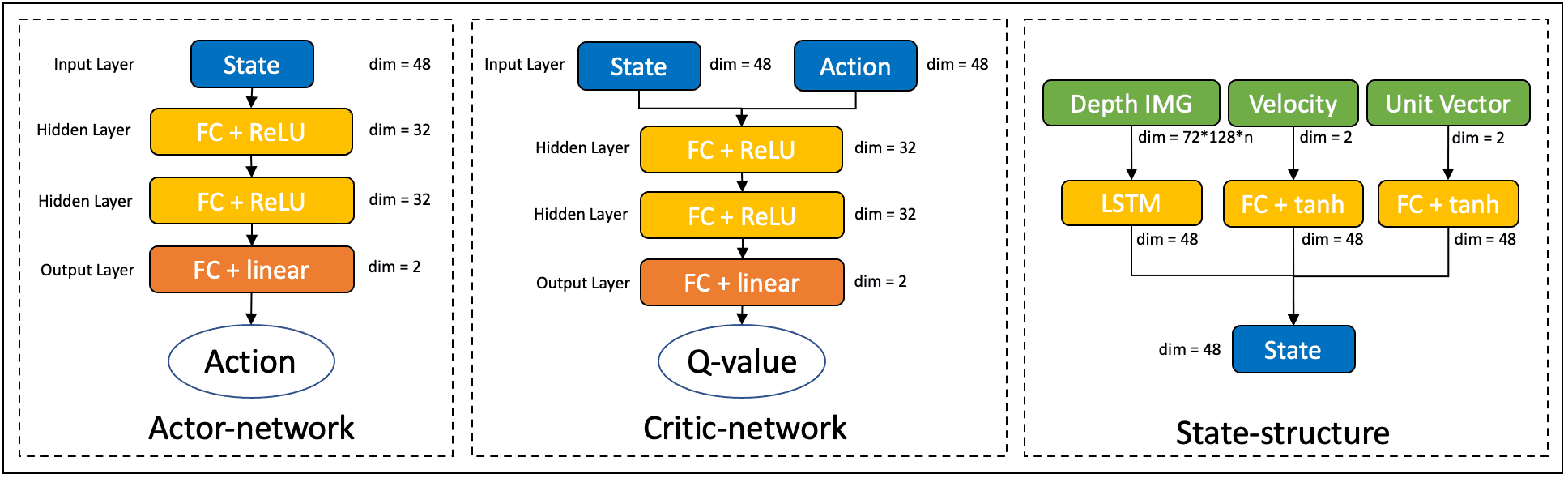}
	\caption{Overall network architecture of our DDPG-based
          method. Left: actor-network; Middle: critic-network;
          Right: information fusion and dimensional reduction through
          LSTM.}
        \label{fig:ddpg}
\end{figure*}

We choose {\it Deep Deterministic Policy Gradient} (DDPG)
\cite{silver2014deterministic} as the DRL algorithm to train the
flight policy of the UAV. DDPG uses an actor-critic method in which
the critic network learns the value function (Q value), and the actor
network decides how the policy model should be updated.
The output of the Actor network can be real-valued vectors, which
enables the DDPG model to directly learn actions in continuous space.
The detailed network structure and state composition of our DDPG
model can be found in Fig.~\ref{fig:ddpg}.
%

Furthermore, for each state, we keep historical information and stack
together depth images from several consecutive time points. This is
designed to alleviate blind spot issues in flight and allow us to
monitor the flight trajectory for smoothness control. To reduce the
dimensionality of the depth map, we use an LSTM network on the depth
stack, as shown in Fig.~\ref{fig:ddpg}, to capture basic
information before feeding it into the actor and critic networks.

\subsection{Reward functions}

In this work, the overall reward $r_t$ at time $t$ is designed to
include multiple components, each of which corresponds to a desired
system condition. 
Note that our design goals include: 1) avoiding collisions, 2)
enhancing model generalization, and 3) encouraging smooth flights.
The overall $r_t$ is given as follows, 

\begin{equation*}
  r_t = R_\textrm{margin} + R_\textrm{towards} + R_\textrm{smooth}  +
  \begin{cases}
    R_\textrm{g} & \text{if at destination} \\ R_\textrm{c} & \text{if
      collision} \\ R_\textrm{f} & \text{if normal flight}
    \end{cases}
    \label{eq:1}
\end{equation*}
where $R_\textrm{margin}$ and $R_\textrm{smooth}$ denote the rewards
to ensure margin and smoothness respectively; $R_\textrm{towards}$ is
aimed to attract the UAV to fly towards the target;
$R_g$ is rewarded if the UAV reaches the end point; $R_c$ is a a
penalty (negative reward) if collisions happen; and $R_f$ contains a
reward for flying forward and a penalty for any deviation from the
predefined route.

\begin{figure}[h]
	\centering
        \includegraphics[width=0.45\textwidth]{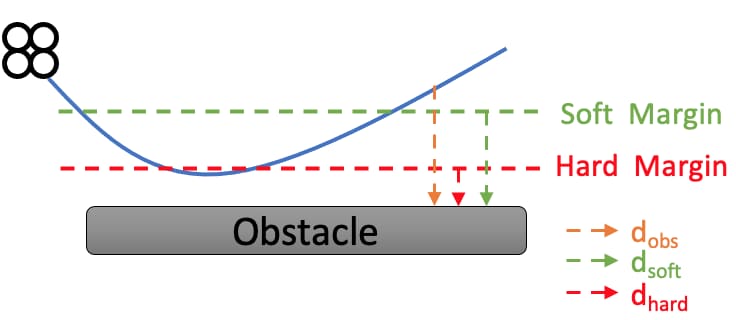}
        \caption{Illustration of the soft margin and hard margin zones.}
	\label{fig:margin}
\end{figure}

$R_\textrm{margin}$ is design to penalize the UAV for getting too
close to the obstacles.
Two margin zones, {\it soft margin} and {\it hard margin} are set up, as
shown in Fig.~\ref{fig:margin}.
When the drone flies into the soft margin zone, it will be pushed back
with a moderate force. If it enters the hard margin zone, the system
should provide a rapidly increasing repulsive force to prevent the
drone from getting closer to the obstacle. Computationally, this
two-margin design is implemented as: 
%
\begin{equation*}
  R_\textrm{margin} =
  \begin{cases}
    -C_1 (d_\textrm{soft} -  d_\textrm{obs}) / ({d_\textrm{soft}} - {d_\textrm{hard}})  & \text{in soft-margin} \\
    -C_2 / {d_\textrm{obs}}  & \text{in hard-margin} \\
    0  & \text{otherwise} 
    \end{cases}
    \label{eq:margin}
\end{equation*}
where $C_1$ and $C_2$ are positive constants; $d_{obs}$ represents the
minimum distance from the drone to the nearest obstacle;
$d_\textrm{soft}$, $d_\textrm{hard}$ are constant parameters denoting
the range of soft zone and hard zone respectively.


$R_\textrm{towards}$ can be designed in many different ways. For
instance, the Euclidean distance away from the destination can be used
as a negative reward to attract the UAV. However, absolute distances
would have poor generalization as they are not scale invariance. In
this work, we use {\it unit vector to the target} as a better
alternative, which is also used as part of the agent state.
Let $v_d$ be the direction (a unit vector) from current position to
the destination and $v_{vel}$ be the direction of the current speed.
We design $R_\textrm{towards} = cos(v_d, v_{vel})$. Comparing with
absolute distances, our design is scale invariant, which leads to two
major benefits: 1) potentially better generalization for unseen
environments; and 2) enhanced training and convergence
performance. The latter comes from the fact that angle to the
destination always provides strong guidance,
even when the UAV is far away from the destination.
%

\subsubsection{Smoothness rewards}

The design of our smoothness rewards is inspired by the classic active
contour model (also known as snakes) proposed by Kass {\it et al}.
for image segmentation \cite{kass1988snakes}.
Let $v(\mathbf{s})$ be an evolving contour. Snake model drives the
contour to capture a target object, with an internal energy term to
ensure the smoothness of the evolution curve:
\begin{equation}
  J(v) = J_1(v) + J_2(v) = \int \alpha |v(\mathbf{s})'|^2 + \beta {|v(\mathbf{s})''|}^2
  \label{J_v} d \mathbf{s}
\end{equation}
where $\alpha$ and $\beta$ are contributing weights. The
Euler-Lagrange (E-L) equations to minimize $J_1(v)$ and $J_2(v)$ are
linear functions and cubic functions, respectively. The former (linear
functions) would force the curve to stretch up into straight line
segments, while the latter (cubic functions) ensures the curve to keep
smooth at all points.

Inspired by the snake model, we take flight trajectories as evolving
curves and design a smoothness reward $R_\textrm{smooth}$ based on the
combination of first-order and second-order derivatives of the
trajectories. More specifically, we intend to minimize the square of
the first-order derivative to stretch up the trajectory, so the UAV
would be forced to fly directly towards the destination. Minimizing
the squared second-order derivative would make the flight path
generally smooth.

Let ${\bf p}_{t}$ denote the location of the UAV at time $t$. The
first-order term will be minimized when ${\bf p}_{t-1}$, ${\bf p}_{t}$,
${\bf p}_{t+1}$ are on the same line.
The difference between $|{\bf p}_{t} - {\bf p}_{t-1}| + |{\bf p}_{t+1}
- {\bf p}_{t}|$ and $|{\bf p}_{t+1} - {\bf p}_{t-1}|$ can be used as
an indictor to measure the deviation of the points from collinearity.
The norm of the second-order derivative can be numerically
approximated with $|{\bf p}_{t+1} + {\bf p}_{t+1} - 2 {\bf p}_{t}|$,
which would be minimized by uniform and gradual changes in direction.
With these two observations, we come up with our smoothness reward at
time $t$:

\begin{equation}
  \begin{split}
    R_\textrm{smooth}  = & 
    -C_3 (|{\bf p}_{t} - {\bf p}_{t-1}| + |{\bf p}_{t+1} - {\bf p}_{t}| - |{\bf p}_{t+1} - {\bf p}_{t-1}|)
    \\
    & -C_4 |\mathbf{p}_{t-1} - 2 \mathbf{p}_{t} + \mathbf{p}_{t+1}|
  \end{split}
  \label{eq:smooth}
\end{equation}
where
$C_3$ and $C_4$ are positive constants, which can be set manually or
empirically in experiments.
It should be noted that, for computational convenience and numerical
stability, we use the norms of the first-order and second-order
derivatives instead of their squared values as in the original snake
model. In addition, our setup of the first-order derivative term,
inspired by the Greedy snake model \cite{williams1992fast}, would
encourage the points to be evenly distributed along the trajectory,
leading to stable flight speed.



\section{Simulation Environments and Experiment Design}

In this work, we train our DRL solutions in one training scene and
test them under three different test scenes, which are previously
unseen. The scenes are designed to evaluate our proposed components
in boosting the generalization capability and trajectory smoothness,
respectively.

\begin{figure}
\centering
  \begin{tabular}{cc}
  \includegraphics[width=4cm]{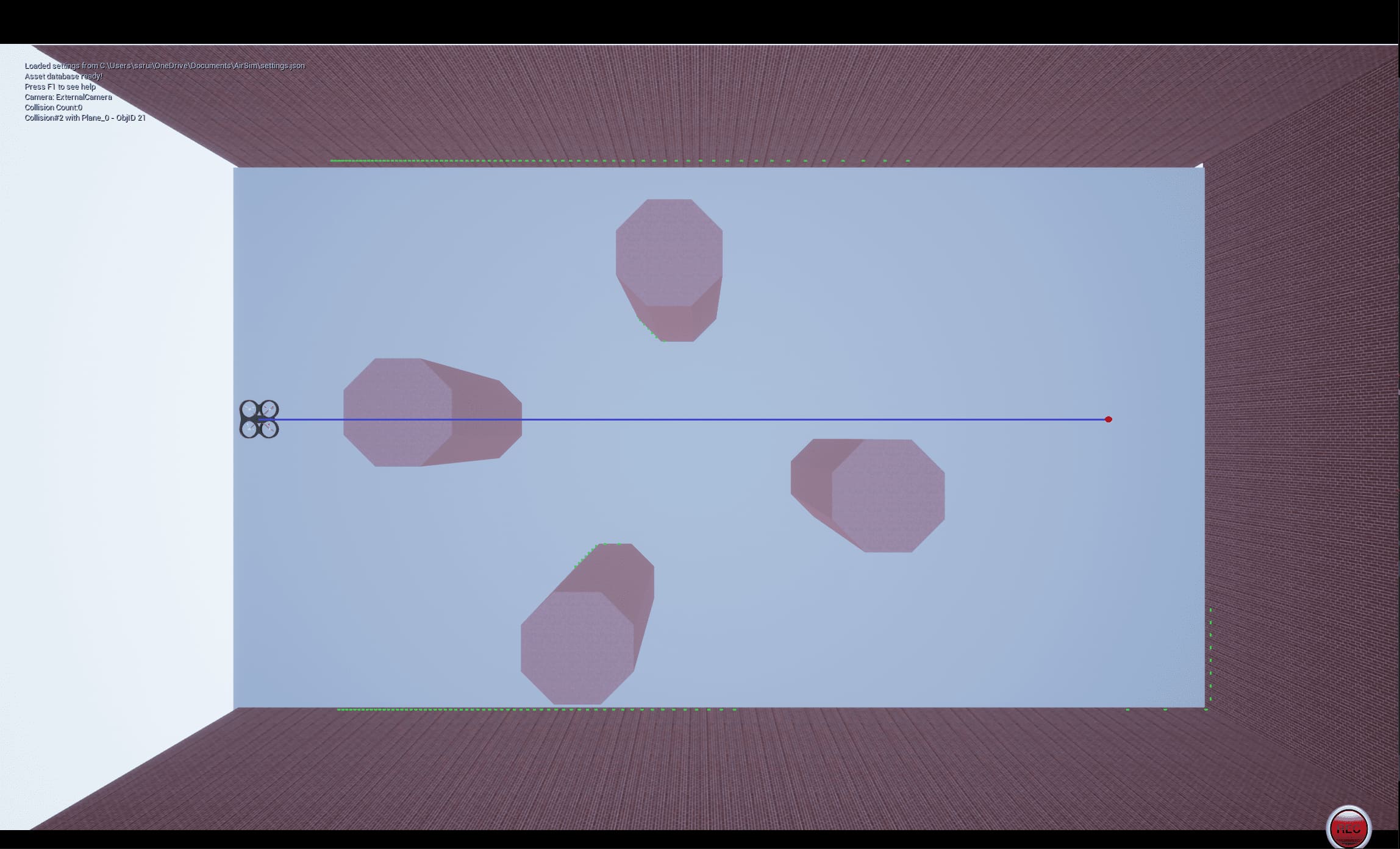} &
  \includegraphics[width=4cm]{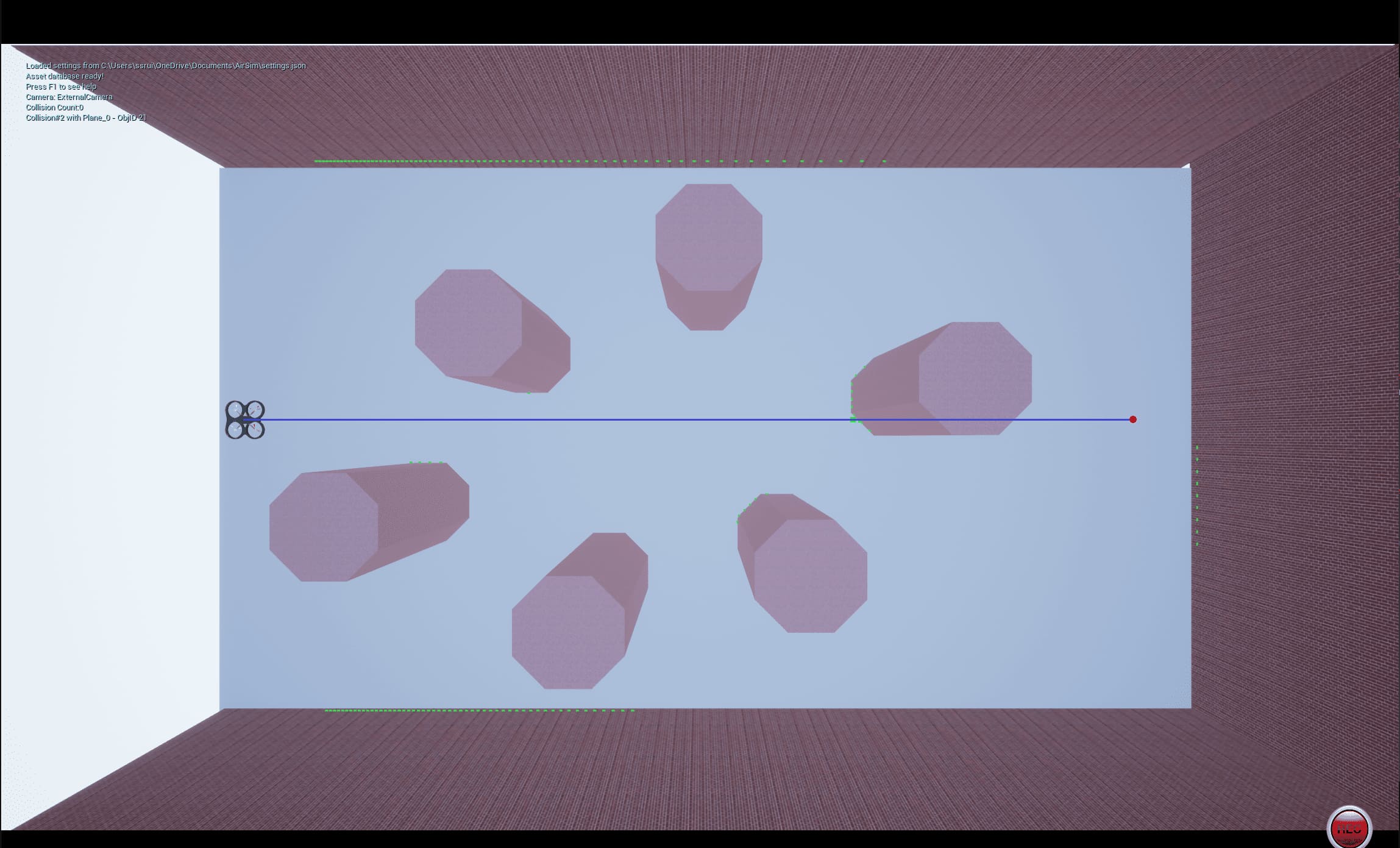} \\ (a):
  Training scene & (b): Test scene 1
  \\ \includegraphics[width=4cm]{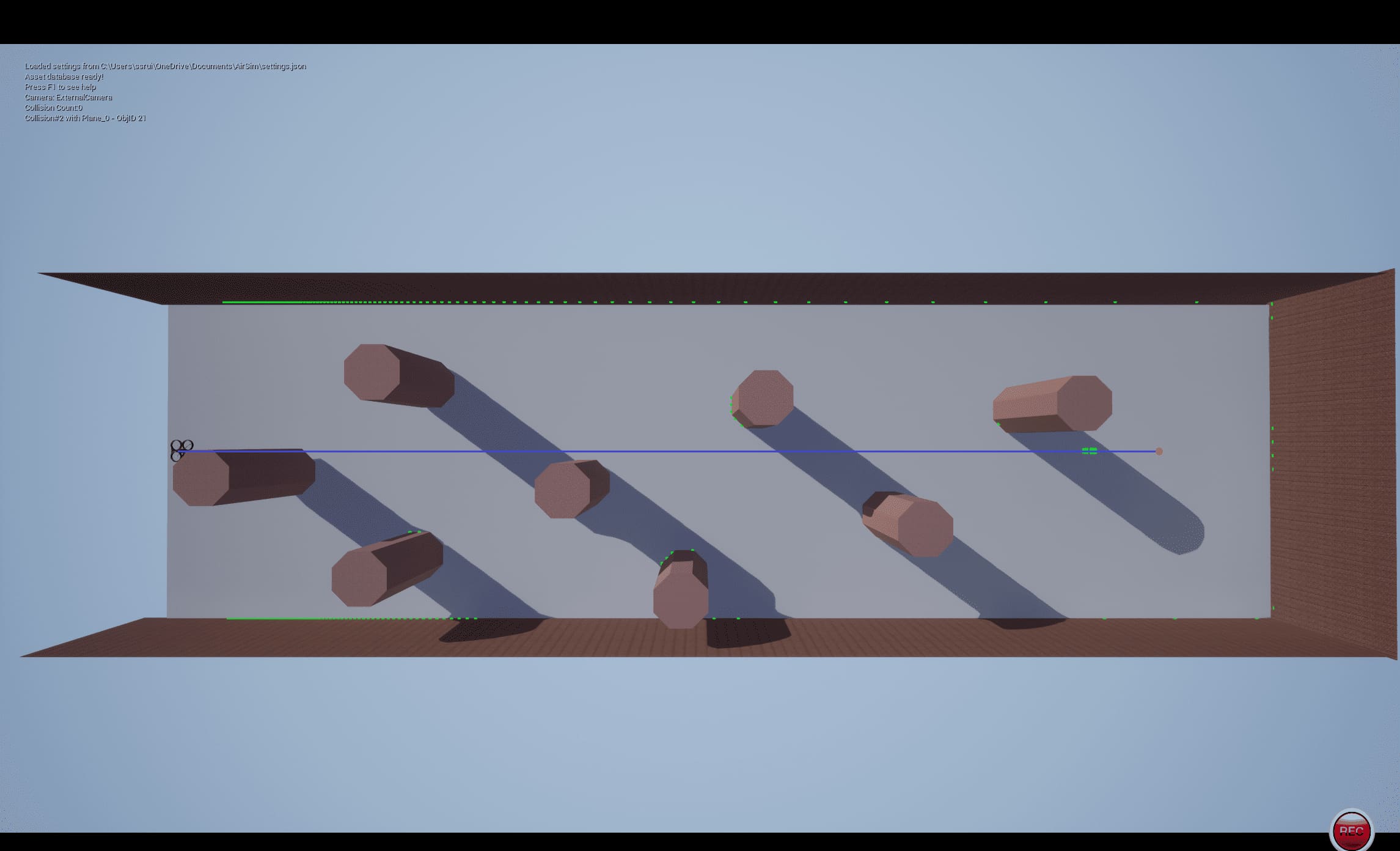} &
  \includegraphics[width=4cm]{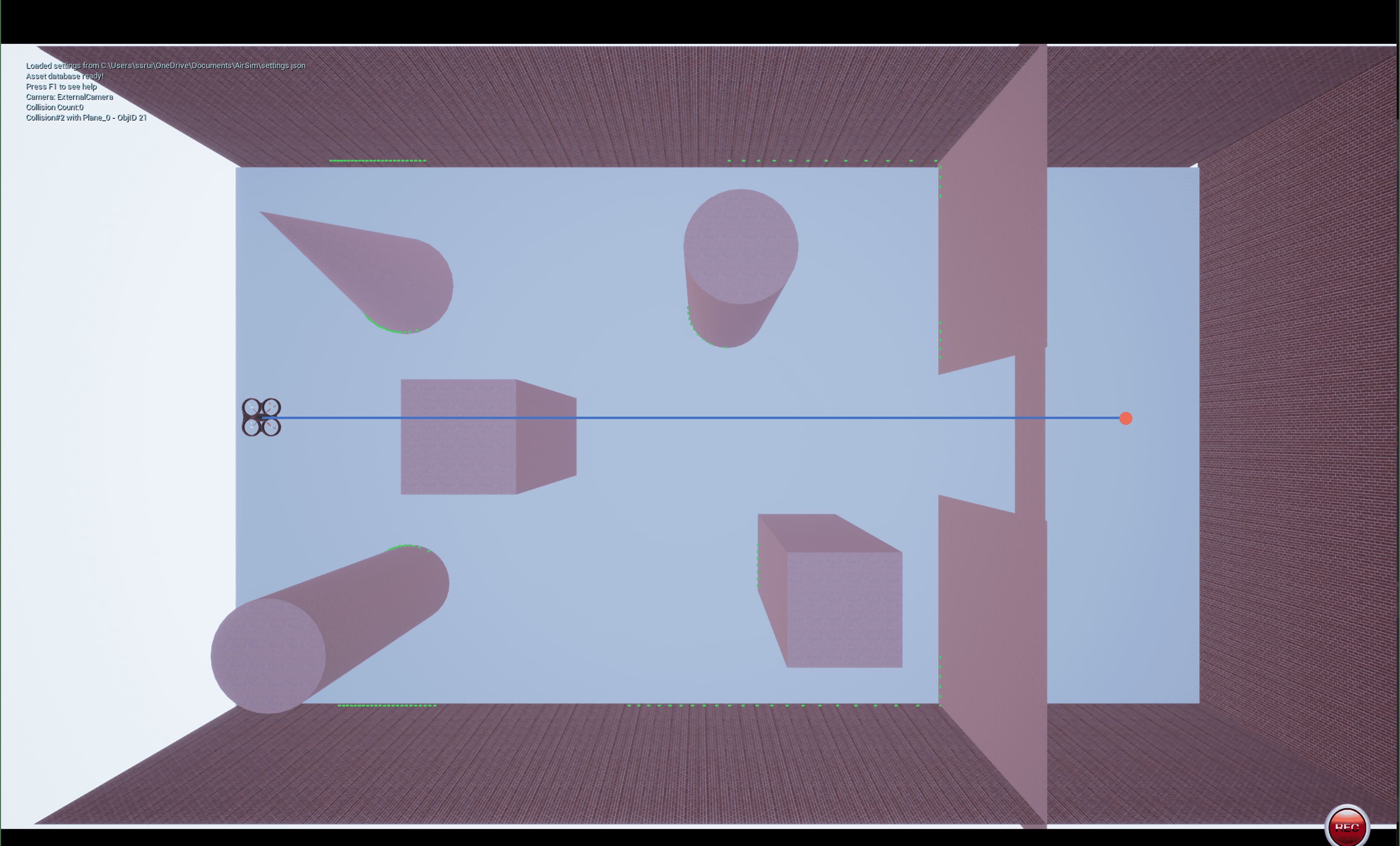} \\ (c): Test
  scene 2 & (d): Test scene 3
\end{tabular}
\caption{Simulation scenes of training and testing environments under
  Airsim + UE4.  (a) is the training scene; (b) has more obstacles;
  (c) is a test scene of larger size and (d) has unseen shapes.  In
  each scene, the solid line is the predefined route and the orange
  point at the right side is the destination. Refer to text for
  details. }
  \label{fig:scenes}
\label{fig:teaser}
\end{figure}


The training scene is designed as a 3D container, 20 meters long, 15
meters high and 15 meters wide.  Four obstacles are put inside, as
shown in Fig.~\ref{fig:scenes}.(a).  The drone takes off at the start
point, and then flies towards the destination, which is marked with an
orange dot. The straight line connecting the two points is the
predefined route. Four thick hexagonal prism (diameter = 2.55m)
obstacles are on the predefined route, blocking the drone flying to
the destination.


Fig.~\ref{fig:scenes}(b - d) show the three test scenes.  The first
test scene (TS1) has the same dimension as the training scene, but
has two additional (totally 6)
hexagonal prism obstacles of the same shape and size. 
This setup
is aimed to test whether the UAV can successfully fly to the end point
in an environment with more crowded obstacles.  The second test scene
(TS2) has eight hexagonal prism obstacles of the same size,
and the total scene length is set to 45 meters.  Compared with the
training scene, this scene is much longer, aiming to evaluate model
generalization under different scene sizes.  The third test scene
(TS3) has the same dimension as the training scene, however, is
filled with obstacles of different shapes.  This is to test the
learned policy in handling obstacle shapes that have not been trained
with.

\section{Experiments and Results}

We build the training and testing simulation scenes on Airsim and UE4,
simulating the UAV collision avoidance in a real environment. The
simulator runs on a Nvidia GeForce RTX 2080 graphics card and the UAV
in the simulator is equipped with a LiDAR, a GPS and an onboard
monocular camera. Experiments are conducted to evaluate the
effectiveness of our designs for 1) boosting the model generalization
and 2) ensuring the smoothness of flight trajectories.



\subsection{Results for smoothness setups}


Our designs for improving the trajectory smoothness comes from the
reward component $R_\textrm{smooth}$ in Eqn.~(\ref{eq:smooth}). For
the convenience of discussion, we use $W_\textrm{smooth}$ to denote
the full DRL model with $R_\textrm{smooth}$ and $WO_\textrm{smooth}$
to denote the model without this term.  Evaluations are conducted based on
visual inspections and statistical analysis, under both the training
and testing environments.

\begin{figure}[h]
\centering
  \begin{tabular}{ccc}
  \includegraphics[width=3.9cm]{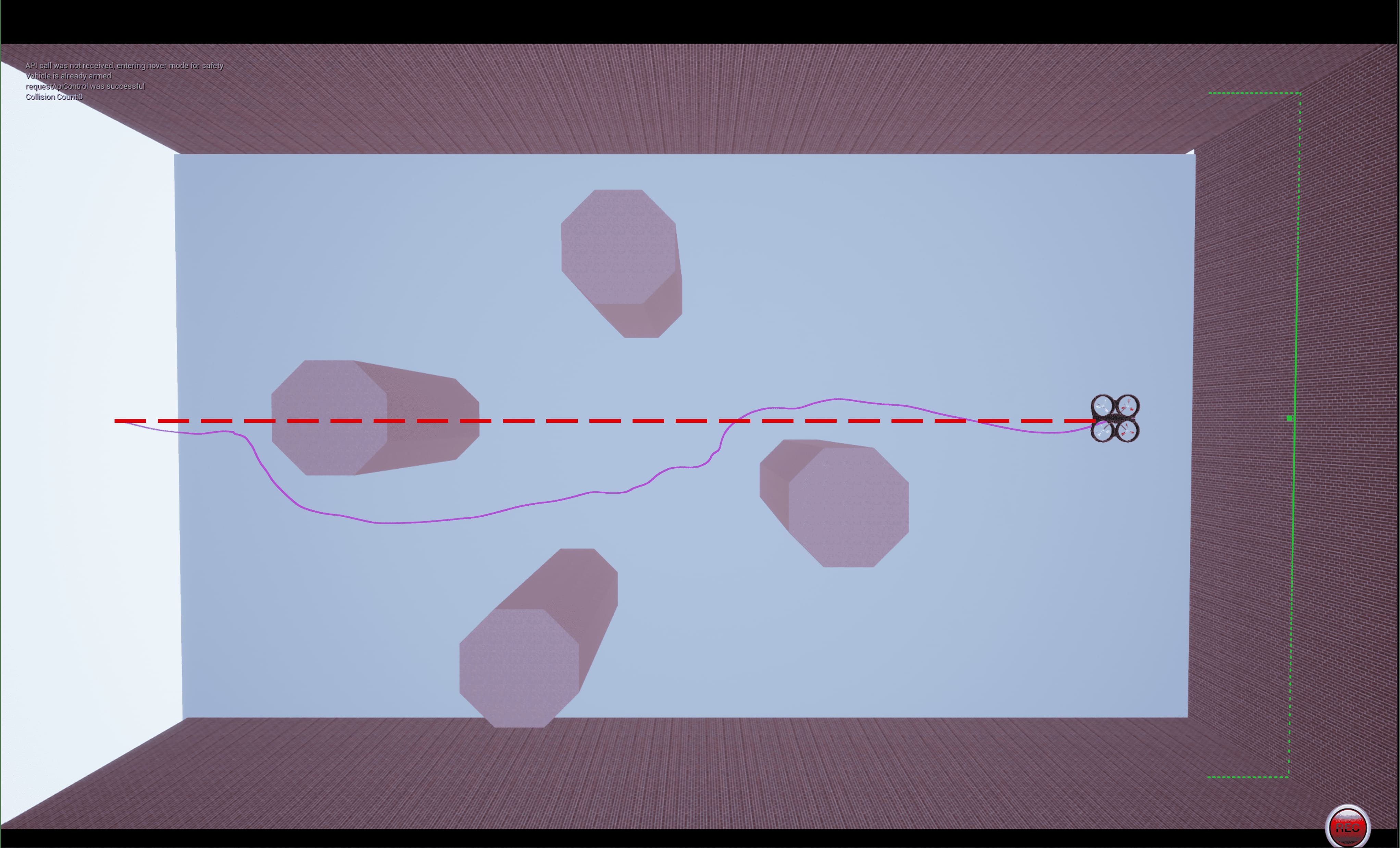} &
  \includegraphics[width=3.9cm]{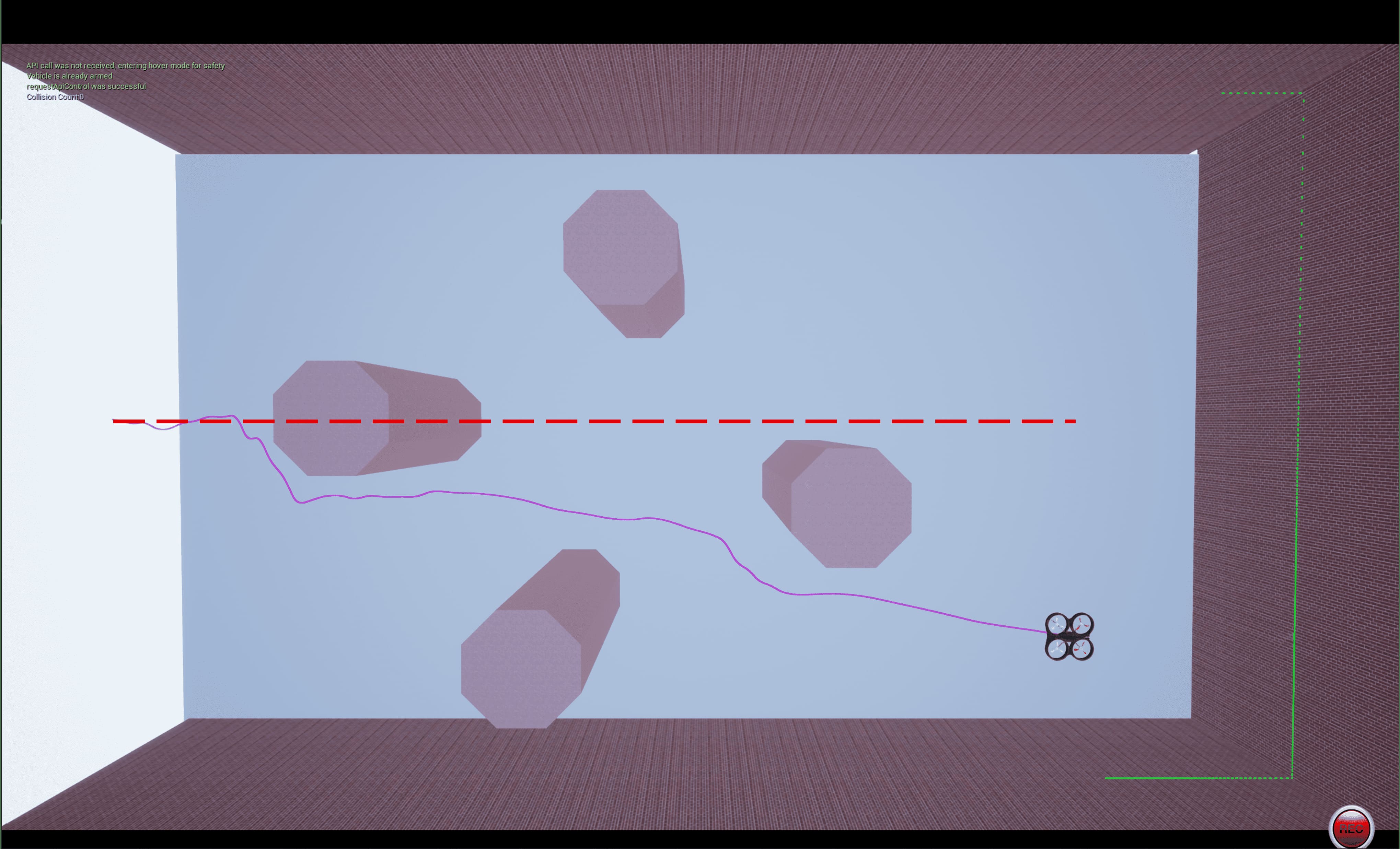} \\
  (a) With
  $R_\textrm{smooth}$ & (b): Without $R_\textrm{smooth}$ 
  \end{tabular}
  \caption{An example pair of training results from the models of (a)
    $W_\textrm{smooth}$ and (b) $WO_\textrm{smooth}$.
  }
  \label{fig:training_stats}
\end{figure}

Fig.~\ref{fig:training_stats} shows an example 
result of the models under the training environment.
Fig.~\ref{fig:training_stats}(a) and Fig.~\ref{fig:training_stats}(b)
record the trajectories of the UAV, 
where the red dashed line is the predefined route.
The trajectory produced by $W_\textrm{smooth}$ is smoother and closer
to the predefined route than that of $WO_\textrm{smooth}$.
%
%


\begin{figure}[h]
\centering
  \begin{tabular}{cc}
  \includegraphics[width=4cm]{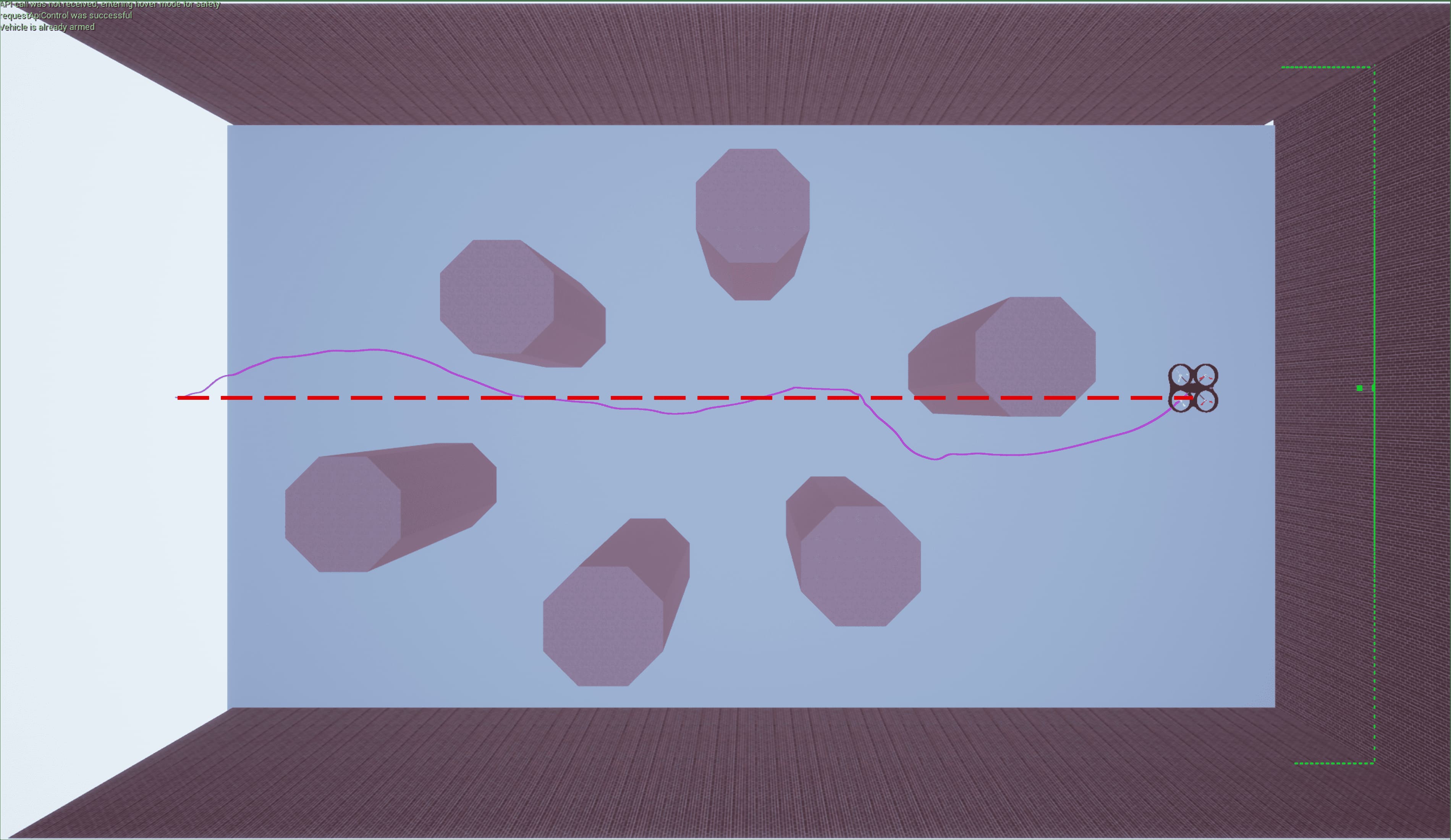} &
  \includegraphics[width=4cm]{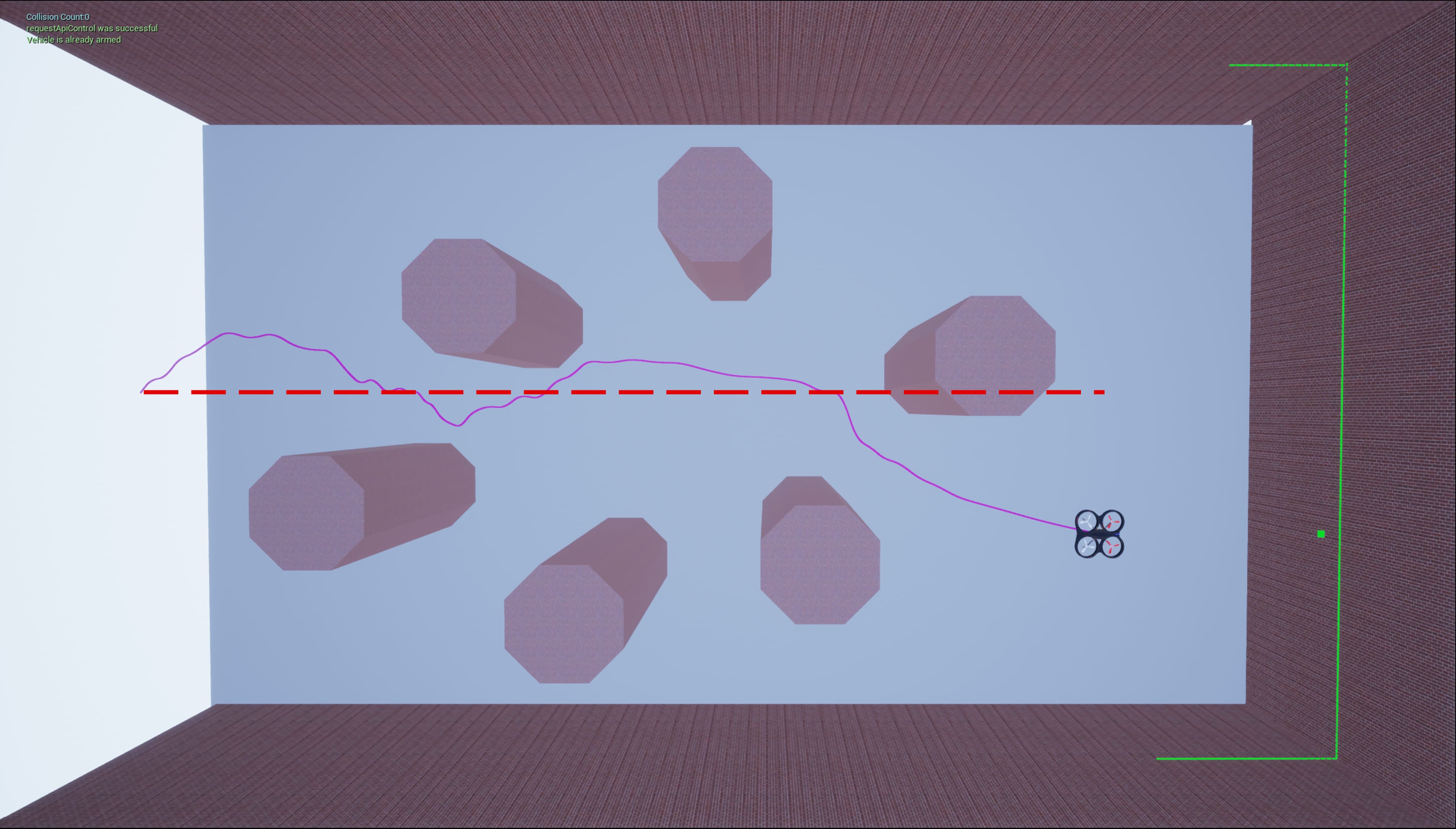}
  \\
  (a): Test 1: with $R_\textrm{smooth}$ & (b): Test 1: without $R_\textrm{smooth}$
    \end{tabular}
  \\
  \begin{tabular}{c}
  \includegraphics[width=7cm]{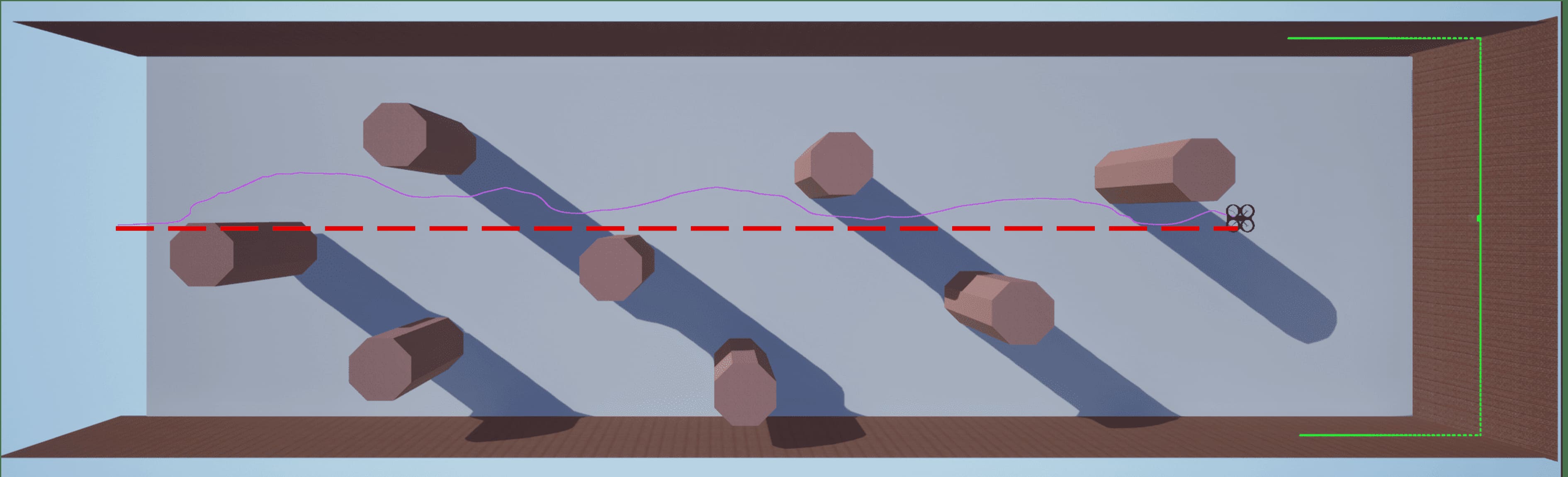}  
  \\
  (c): Test 2: with   $R_\textrm{smooth}$
  \\
  \includegraphics[width=7cm]{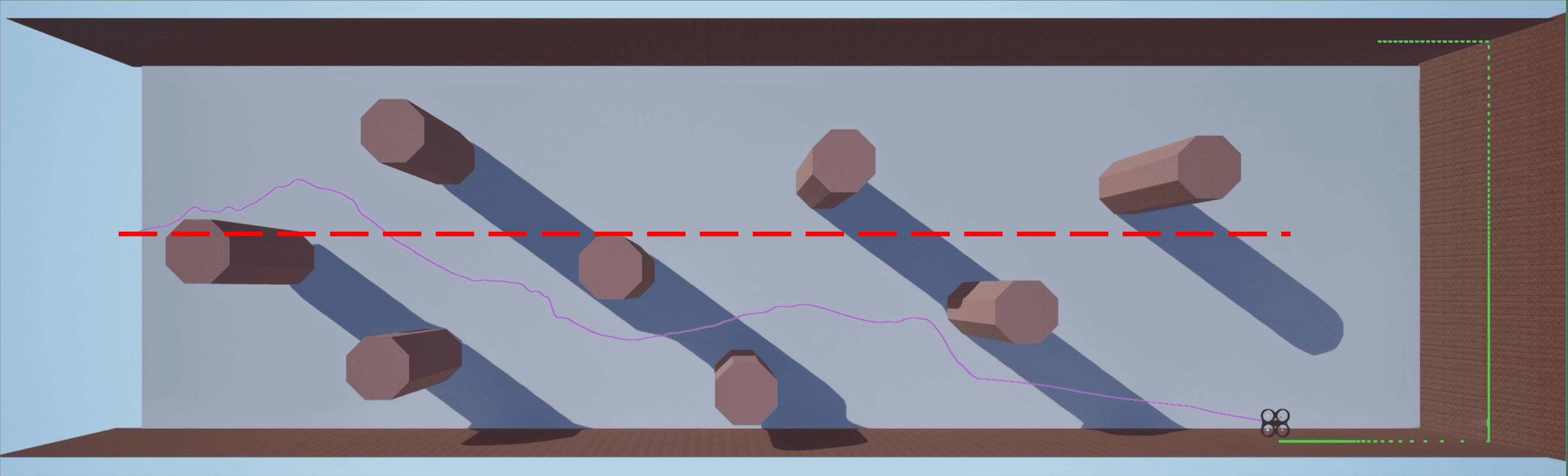}
  \\ (d): Test 2: without $R_\textrm{smooth}$
  \end{tabular}
  \\
  \begin{tabular}{cc}
  \includegraphics[width=4cm]{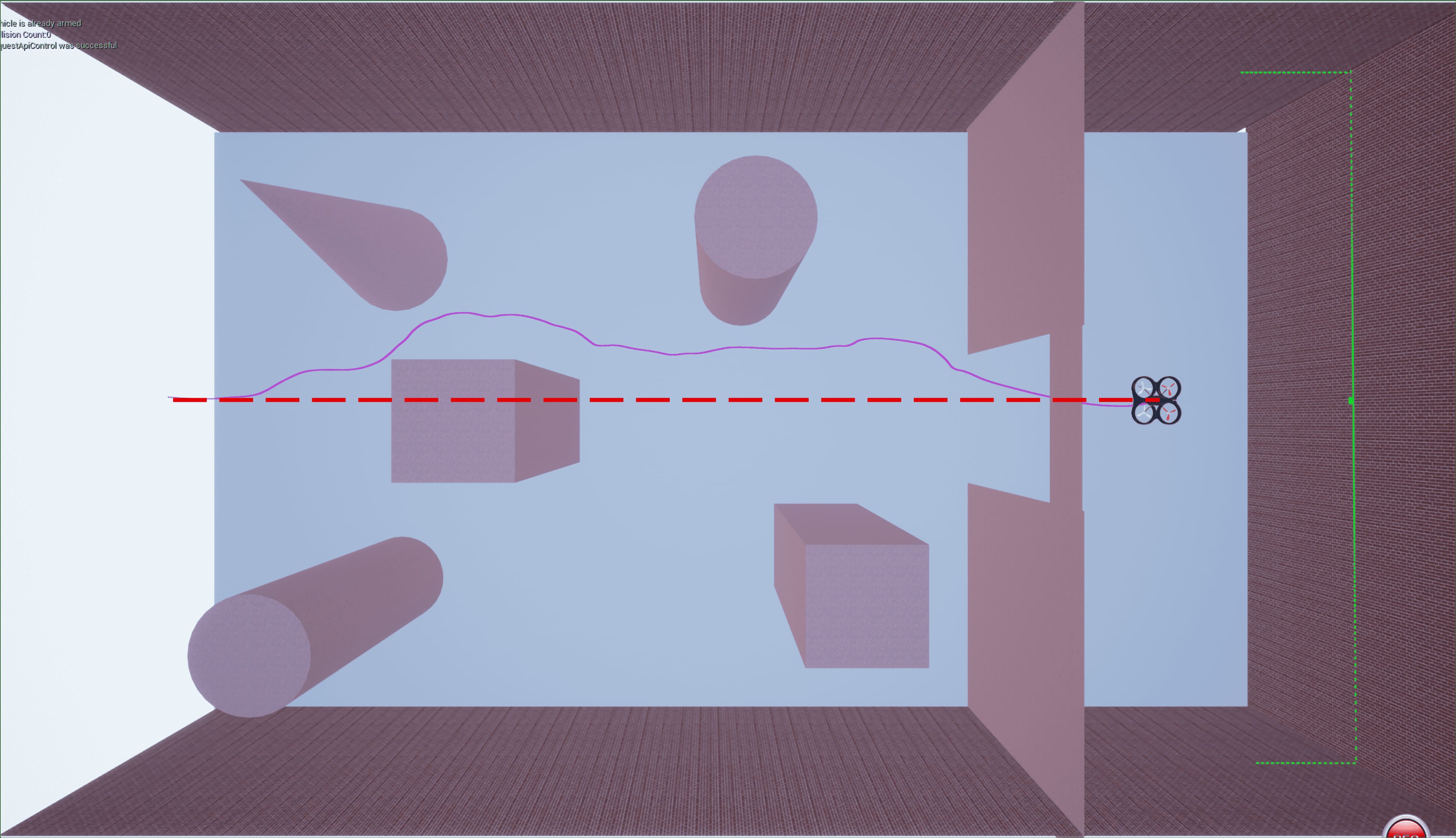} & 
  \includegraphics[width=4cm]{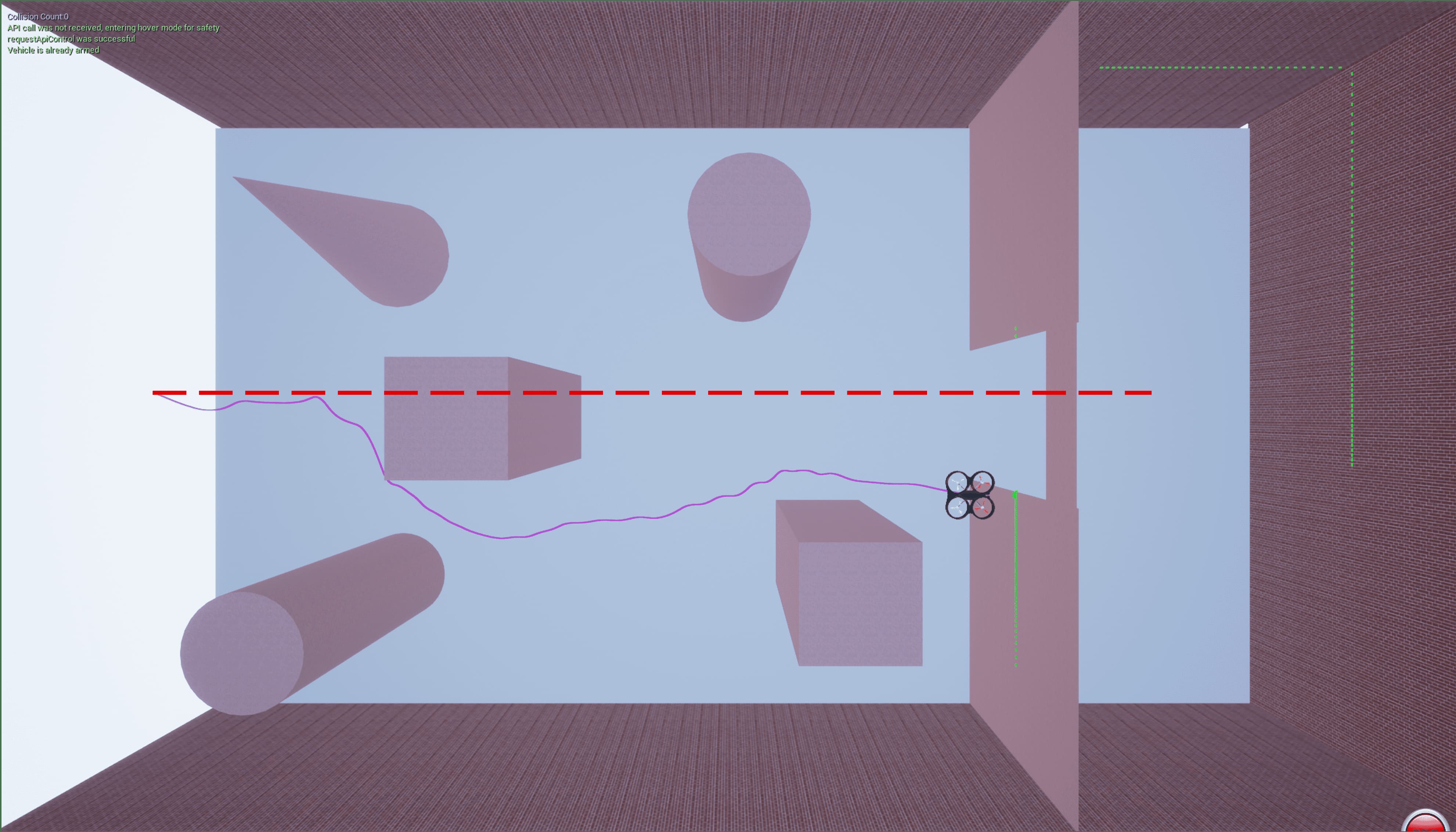}
  \\
  (e): Test 3:
    with $R_\textrm{smooth}$ & (f): Test 3: without
    $R_\textrm{smooth}$
  \end{tabular}
  
  \caption{(a) and (b): final trajectories of $W_\textrm{smooth}$ and
    $WO_\textrm{smooth}$ in Test scene 1.  (c) and (d): final
    trajectories of the two models in Test scene 2.  (e) and (f):
    final trajectories of the two models in Test scene 3. }
  \label{fig:test_tracks}
\end{figure}

For the inference performance in the unseen test environments,
Fig.~\ref{fig:test_tracks} shows an example of the trajectories
generated by the two models.
In each figure, the dashed line is the predefined route and the purple
line shows the trajectory of the UAV.
One can see $W_\textrm{smooth}$ clearly outperforms
$WO_\textrm{smooth}$ in all three test scenes, producing smoother and
more direct tracks towards the destination. In the third test scene,
$WO_\textrm{smooth}$ can not
find the path to the destination, as the UAV collides with the cube
and the door, indicating that it fails to bypass obstacles of unseen
shapes. In contrast, $W_\textrm{smooth}$, with the proposed smoothness
rewards, can successfully find a rather smooth path to reach the
destination.

We conduct multiple inference tests of the two models under the three
test scenes. The numbers of the tests are 100, 50 and 100,
respectively, for test scenes 1, 2 and 3. Three evaluation metrics are
employed to evaluate the policy performance: {\it average linear
  acceleration} over the flights, {\it average path curvature} of the
flight trajectories, and {\it success rate} to reach the
destinations. The comparisons are summarized in Table~\ref{test_data}.
%
%
%
Our $W_\textrm{smooth}$ model has lower average curvatures and
accelerations than $WO_\textrm{smooth}$, demonstrating
that the produced trajectories are generally smoother and more stable.
Our model also has higher success rates to reach destinations,
indicating the smooth terms also make positive contributions to
training convergence.

It should be noted that our setup of the first-order derivative in the
smoothness term in Eqn.~(\ref{eq:smooth}) not only encourages the
flight trajectory to be tighten up, but also drives the points to be
distributed evenly \cite{williams1992fast}. This would lead to more
stable flight speed, which is partly reflected in the smaller values
of the average acceleration. Both first-order and second-order
derivatives contribute to the overall smoothness. The smaller average
curvatures produced by $W_\textrm{smooth}$ indicate we have achieved
our design goals.


\begin{table*}[h]
  \centering
  \caption{Quantitative results for $WO_\textrm{smooth}$ and
    $W_\textrm{smooth}$ under the unseen test scenes. Acc. is short
    for {\it Acceleration}; Cur. is short for {\it Curvature} and
    Succ.  is for {\it Success Rate}. }
  \label{test_data}
  \scalebox{1.05} 
           {
	     \begin{tabular}{|c|c|c|c|c|c|c|}
               \hline \hline \multirow{2}*{Scene} &
               \multicolumn{3}{c|}{ $WO_\textrm{smooth}$} &
               \multicolumn{3}{c|}{$W_\textrm{smooth}$}
               \\ \cline{2-7} & Acc. ($m/s^2$) & Cur. ($m^{-1}$) &
               Succ. (\%) & Acc. ($m/s^2$) & Cur. ($m^{-1}$) &
               Succ. (\%) \\ \hline Test scene 1 & 0.07 & 0.08 & 99 &
               0.03 & 0.06 & 90 \\ \hline Test scene 2 & 0.04 & 0.08 &
               45 & 0.03 & 0.05 & 57 \\ \hline Test scene 3 & NA & NA
               & 0 & 0.04 & 0.06 & 100 \\ \hline
	     \end{tabular}
           }
\end{table*}

%

\subsection{Results for model generalization}


In this work, we propose two practical designs for improving the DRL
agent's handling of environment changes: 1) using {\it shallow depth}
instead of {\it deep depth} as a part of agent's state; and 2) using
{\it unit vector towards destination} (denoted as $\vv{vec}$) instead
of the distance (denoted as $dist$) in agent's state and reward
functions.

In order to test the effectiveness of our designs, we design four DRL
models based on: 1) {\it deep depth} + $\vv{vec}$; 2) {\it shallow
  depth} + $\vv{vec}$; 3) {\it deep depth} + $dist$ and 4) {\it
  shallow depth} + $dist$. It should be noted that $\vv{vec}$ is used
in both the state and reward of our models. In models 3) and 4), both
$\vv{vec}$ terms are replaced with $dist$.  To simplify the analysis,
smoothness reward is not integrated into the models.
For each of the combination, we train the policy under the training
environment to convergence first, and then directly perform inferences
in the three unseen test environments described in the previous
section. Note that test scene TS1 has more crowded obstacles than the
training scene; test scene TS2 has a larger size; and test scene TS3
is filled with objects that have not been seen in the training.






The four models are all trained successfully in the training
scene. When moving into the three testing environments, we conducted
100 trials for each of the TS1, TS2 and TS3 environments.

To measure the performance of the trained policies in the unseen
testing environments, we design two evaluation metrics: 1) {\it
  success rate} (SR), which is the percentage of the trials where the
UAV can successfully arrive at the destination; and 2) {\it capability
  of collision avoidance} (CAC), which is defined as the average
length that the UAV can fly prior to its first collision.  The results
are shown in Table \ref{table:2}.

\begin{table}[h]
  \centering
  \caption{Comparisons of the 4 policies in the unseen testing
    environments, measured with {\it success rate} (SR) and {\it
      capability of collision avoidance (CAC).}}
  \label{table:2}
  \scalebox{1} 
           {
             \begin{tabular}{|c|c|c|c|c|}
               \hline \hline  & {\it deep + dist} & {\it shallow + dist}  & {\it deep} + $\vv{vec}$  & {\it shallow} + $\vv{vec}$                \\
               \hline 
               SR  & 3 $\%$ & 18.33$\%$ & 3 $\%$ & 63.67$\%$ \\
               \hline
               CAC  & 50.18$\%$ & 55.10$\%$  & 57.25$\%$ & 87.33$\%$ \\
               \hline
             \end{tabular}
           }
\end{table}

Comparing with the other three models, {\it shallow} + $\vv{vec}$
demonstrates a much stronger capability of finding the targets in
unseen environments. It also shows the best capability in avoiding
collisions.  {\it deep depth} + $dist$, on the other hand, show the
worst performance in CAC.

From the above experiments, we can observe that both {\it shallow
  depth} and {\it unit vector} are helpful for the learning policy to
be directly deployed in changed environments. Their combination
demonstrates the best generalization capability in the experiments.

 \section{Conclusions}

In this work, we propose a DRL-based collision avoidance solution for
UAVs to fly smoothly and stably. We also seek to develop our system to
have good generalization performance to handle unseen environments. A
number of novel designs are made regarding the agent’s state, as well
as its reward function. The major innovation is a smoothness reward
term, based on the minimization of a combined first-order and
second-order derivatives. Experiments demonstrate this term can indeed
lead to smoother trajectories, as well as flights with stable speeds.

Our design for boosting model generalization is based on shallow depth
and unit vector towards target. While simple, this combined setup
reduces the input space and increases system invariance, leading to
greatly enhanced robustness. Experimental results demonstrate the
effectiveness of our overall design and individual components.  To
transplant proposed solutions and deploy them into real UAVs is our
ongoing effort.



\large
\bibliographystyle{IEEEtran}
\bibliography{citation}

\end{document}